\documentclass[twoside]{article}

\usepackage[accepted]{aistats2022}
\usepackage{apacite}
\newcommand{\citet}[1]{\citeauthor{#1} \shortcite{#1}}
\newcommand{\citep}{\cite}

\usepackage{pgfplots}
\usepackage{calc}
\usepackage{times}
\usepackage{epsfig}
\usepackage{graphicx}
\usepackage{amsmath}
\usepackage{amssymb}
\usepackage{bm}
\usepackage{algorithm}
\usepackage{algpseudocode}
\usepackage{booktabs}
\usepackage{units}
\usepackage{caption}
\usepackage{subcaption}
\usepackage{tabularx}

\def\eqref#1{equation~\ref{#1}}

\def\1{\bm{1}}

\def\vtheta{{\bm{\theta}}}

\def\vphi{{\bm{\phi}}}

\def\va{{\bm{a}}}
\def\vb{{\bm{b}}}

\def\vg{{\bm{g}}}

\def\vr{{\bm{r}}}

\def\vx{{\bm{x}}}
\def\vy{{\bm{y}}}
\def\vz{{\bm{z}}}

\DeclareMathAlphabet{\mathsfit}{\encodingdefault}{\sfdefault}{m}{sl}
\SetMathAlphabet{\mathsfit}{bold}{\encodingdefault}{\sfdefault}{bx}{n}

\def\gX{{\mathcal{X}}}
\def\gY{{\mathcal{Y}}}

\def\DsTrain{{\mathcal{D}^{\text{train}}}}
\def\DsTraini{{\mathcal{D}_i^{\text{train}}}}
\def\Task{{\mathcal{T}}}

\begin{document}

%

%

\twocolumn[

\aistatstitle{MT3: Meta Test-Time Training for Self-Supervised Test-Time Adaption}

\aistatsauthor{ Alexander Bartler \And Andre B\"uhler \And Felix Wiewel \And Mario D\"obler \And Bin Yang }

\aistatsaddress{Institute of Signal Processing and System Theory, University of Stuttgart, Germany } ]

\begin{abstract}
An unresolved problem in Deep Learning is the ability of neural networks to cope with domain shifts during test-time, imposed by commonly fixing network parameters after training. Our proposed method Meta Test-Time Training (MT3), however, breaks this paradigm and enables adaption at test-time. We combine meta-learning, self-supervision and test-time training to learn to adapt to unseen test distributions. By minimizing the self-supervised loss, we learn task-specific model parameters for different tasks. A meta-model is optimized such that its adaption to the different task-specific models leads to higher performance on those tasks. During test-time a single unlabeled image is sufficient to adapt the meta-model parameters. This is achieved by minimizing only the self-supervised loss component resulting in a better prediction for that image. Our approach significantly improves the state-of-the-art results on the CIFAR-10-Corrupted image classification benchmark. Our implementation is available on GitHub$^1$.

\end{abstract}

\section{Introduction}
\begin{figure}[!t]
	\centering
	\begin{subfigure}[b]{0.18\linewidth}
		\centering
		\includegraphics[width=\linewidth]{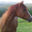}
		\caption{original}
	\end{subfigure}~
	\begin{subfigure}[b]{0.18\linewidth}
		\centering
		\includegraphics[width=\linewidth]{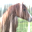}
		\caption{snow}
	\end{subfigure}~
	\begin{subfigure}[b]{0.18\linewidth}
		\centering
		\includegraphics[width=\linewidth]{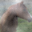}
		\caption{fog}
	\end{subfigure}~
	\begin{subfigure}[b]{0.18\linewidth}
		\centering
		\includegraphics[width=\linewidth]{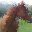}
		\caption{glass}
	\end{subfigure}~
	\begin{subfigure}[b]{0.18\linewidth}
		\centering
		\includegraphics[width=\linewidth]{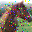}
		\caption{impulse}
	\end{subfigure}~
	\caption{Example of a CIFAR-10 image with different corrupted versions of the most severe level taken from the CIFAR-10-Corrupted dataset.}
	\label{fig:corr}
\end{figure}

Deep neural networks have dramatically improved the results in a wide range of applications. However, after they are deployed the distribution of test data may be very different compared to the distribution of the training data. During testing, samples may be corrupted by, e.g., noise, different lighting conditions, or environmental changes such as snow or fog (see Figure \ref{fig:corr}). These corruptions and the resulting distribution shifts can cause a dramatic drop in performance \cite{Azulay2019, hendrycks2019benchmarking}. Even truly unseen test images without a large distribution shift can harm the model performance \cite{recht19aCifar101}. Adversarial perturbations are examples for an intentional distribution shift, that is not recognizable to humans, but also reduces model performance drastically. 

To address changes in the test distribution and the resulting performance drop, recent work has mainly focused on the robustness to adversarial examples \cite{carlini2017adversarial,Chen_2020_CVPR,Dong_2020_CVPR,Jeddi_2020_CVPR,szegedy2013intriguing} or the generalization to out-of-distribution samples \cite{albuquerque2020improving,hendrycks2020many,krueger2020out}. Both areas aim to train a model in order to be robust against various types of unknown corruptions, distribution shifts, or domain shifts during testing. Another concept  assumes that during training multiple unlabeled samples of the target domain are available and therefore unsupervised domain adaptation (UDA) can be performed \cite{tan2018survey,wang2018deep, wilson2020survey,zhao2020review}. In the extreme case of one-shot UDA, only one unlabeled sample of the target domain is available during training \cite{luo2020adversarial,Benaim2018OneShotUC}.

On the contrary, it is possible to account for distribution shifts only during test-time using a single test image under the assumption that the test image contains information about the distribution it originates from. Since the adaption to a single test sample is then performed by adapting the model at test-time, there is no need for any test data or information about the test distribution during the training stage. Additionally, in contrast to UDA, neither the original training data is needed for the adaption to a new test sample nor the test samples have to be drawn from the same distribution, since each test sample is processed individually. The assumption that each test sample can be corrupted differently could occur more likely in practice than having a persistent shift of the test distribution after deployment. For the concept of adaption during test-time, the model can be quickly adapted using only the sample itself, where in one-shot UDA the complete training dataset is additionally used to train a model on the target domain. If the target distribution is not stationary, the one-shot UDA has to be applied for each test sample individually, which would result in a tremendous testing complexity.

\paragraph{Test-Time Training}
The concept of adaption during test-time was first proposed by \citeA{sun2020ttt} and is called Test-Time Training (TTT). In order to train a model which is able to adapt to unseen images, \citeA{sun2020ttt} proposed to train the model using a supervised and a self-supervised loss jointly, denoted as joint training. During testing, only the self-supervised loss on one unlabeled test image is applied to adapt the model parameters which is hence called test-time training. After that, the refined model is used for inference. This test-time adaption is done for each test sample individually starting from the initially trained model.

The used architecture has two heads and a shared feature extractor. One head is used for the supervised downstream task, e.g., for classification the minimization of the categorical cross-entropy loss. The second head enables self-supervised learning. It solves a simple auxiliary task of rotation prediction, where four different rotation angles have to be predicted in a four-way classification problem \cite{gidaris2018rot}. 

During testing, a batch of augmented views of a single image is used to minimize only the self-supervised loss subsequently to adapt the shared feature extractor while keeping the head for the supervised downstream task unchanged. The adapted model is used to make the prediction of the test sample. In addition, \citeA{sun2020ttt} showed under strong assumptions that minimizing the self-supervised loss during testing implicitly minimizes the supervised loss. 

While for the standard test procedure the adapted model is only used for a single image, the authors proposed an online setting where the model parameters are adapted sequentially during testing with a stream of test samples of a stationary or gradually changing test distribution. This online setting can be seen as online unsupervised domain adaption. Another recent approach for online test-time adaption is built upon entropy minimization \cite{wang2020tent}. 

\paragraph{Self-Supervised Learning}
In the work of \citeA{sun2020ttt} the rather simple auxiliary task of rotation prediction \cite{gidaris2018rot} is used for self-supervision.  Recent state-of-the-art approaches for representation learning, on the other hand, rely on contrastive learning \cite{Chen2020Simclr,Chen2020_simclrv2,he2020momentum,Oord2018cpcv1}. The key idea of contrastive learning is to jointly maximize the similarity of representations of augmented views of the same image while minimizing the similarity of representations of other samples, so called negatives. Another state-of-the-art technique for self-supervised representation learning  is called \textit{Bootstrap Your Own Latent} (BYOL) \cite{grill2020bootstrap}. Compared to contrastive losses, the main advantage of BYOL is that there is no need for negative samples. This makes BYOL suitable for test-time training since  there is only a single image available during test-time.   

BYOL consists of two neural networks, the online and target model. Both networks predict a representation of two different augmented views of the same image. The online network is optimized such that both the online and target predictions of the two augmented views are as similar as possible. This is realized by minimizing the mean squared euclidean distance of both $l_2$-normalized predictions. The parameters of the target network are updated simultaneously using an exponential moving average of the online network parameters.  

\begin{figure*}[!th]
	\centering
	\begin{subfigure}[b]{0.45\linewidth}
		\begin{center}
\begingroup%
  \makeatletter%
  \providecommand\color[2][]{%
    \errmessage{(Inkscape) Color is used for the text in Inkscape, but the package 'color.sty' is not loaded}%
    \renewcommand\color[2][]{}%
  }%
  \providecommand\transparent[1]{%
    \errmessage{(Inkscape) Transparency is used (non-zero) for the text in Inkscape, but the package 'transparent.sty' is not loaded}%
    \renewcommand\transparent[1]{}%
  }%
  \providecommand\rotatebox[2]{#2}%
  \newcommand*\fsize{\dimexpr\f@size pt\relax}%
  \newcommand*\lineheight[1]{\fontsize{\fsize}{#1\fsize}\selectfont}%
  \ifx\svgwidth\undefined%
    \setlength{\unitlength}{226.77165354bp}%
    \ifx\svgscale\undefined%
      \relax%
    \else%
      \setlength{\unitlength}{\unitlength * \real{\svgscale}}%
    \fi%
  \else%
    \setlength{\unitlength}{\svgwidth}%
  \fi%
  \global\let\svgwidth\undefined%
  \global\let\svgscale\undefined%
  \makeatother%
  \begin{picture}(1,0.5)%
    \lineheight{1}%
    \setlength\tabcolsep{0pt}%
    \put(0,0){\includegraphics[width=\unitlength]{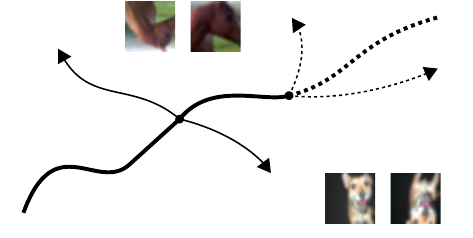}}%
    \put(0.02341076,0.43372612){\color[rgb]{0,0,0}\makebox(0,0)[lt]{\lineheight{1.25}\smash{\begin{tabular}[t]{l}$\nabla_{\phi_1}\mathcal{L}_{\text{BYOL}}\bigg(\qquad \quad ,\qquad \quad \bigg)$\end{tabular}}}}%
    \put(0.44012938,0.06992416){\color[rgb]{0,0,0}\makebox(0,0)[lt]{\lineheight{1.25}\smash{\begin{tabular}[t]{l}$\nabla_{\vphi_2}\mathcal{L}_{\text{BYOL}}\bigg(\qquad \quad ,\qquad \quad \bigg)$\end{tabular}}}}%
    \put(0.54608226,0.24166374){\color[rgb]{0,0,0}\makebox(0,0)[lt]{\lineheight{1.25}\smash{\begin{tabular}[t]{l}$\nabla_{\vtheta}\mathcal{L}_{\text{total}}$\end{tabular}}}}%
    \put(0.19145323,0.16686058){\color[rgb]{0,0,0}\makebox(0,0)[lt]{\lineheight{1.25}\smash{\begin{tabular}[t]{l}$\vtheta$\end{tabular}}}}%
    \put(0.43952084,0.17319374){\color[rgb]{0,0,0}\makebox(0,0)[lt]{\lineheight{1.25}\smash{\begin{tabular}[t]{l}$\vphi_2$\end{tabular}}}}%
    \put(0.25171394,0.32225761){\color[rgb]{0,0,0}\makebox(0,0)[lt]{\lineheight{1.25}\smash{\begin{tabular}[t]{l}$\vphi_1$\end{tabular}}}}%
  \end{picture}%
\endgroup%

			\label{subfig:training}
		\end{center}
		\caption{meta-training}
	\end{subfigure}~
	\begin{subfigure}[b]{0.45\linewidth}
		\begin{center}
			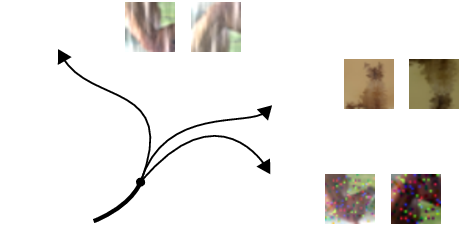
		\end{center}
		\caption{test-time adaption}
		\label{subfig:test}
	\end{subfigure}~
	\caption{MT3 training and test-time adaption: (a) In the outer loop the meta-parameters $\vtheta$ are updated to minimize $\mathcal{L}_{\text{total}}$ which  depends on multiple inner loop adaption steps. In the inner loop, different augmented views of the same image are used to adapt the meta-model to the task-specific model. (b) Test-time adaption to different corruptions starting from optimized meta-parameters $\vtheta^*$. }
	\label{fig:principle}
\end{figure*}
\paragraph{Meta-Learning}
Another concept of adapting to unknown tasks or distributions is meta-learning, which is used in many state-of-the art results, e.g., for supervised few-shot learning \cite{antoniou2018train,finn2017maml,hospedales2020meta,li2017meta,nichol2018first} or unsupervised few-short learning \cite{hsu2018unsupervised,khodadadeh2018unsupervised}.  Meta-learning has also shown its flexibility in the work of \citeA{metz2018meta} where an unsupervised update rule is learned which can be used for pre-training a network in order to get powerful representations of unknown data distributions. In the work of \citeA{Balaji2018metareg}, meta-learning is used to train models that generalize well to unknown domains. A widely used optimization based meta-learning algorithm is Model-Agnostic Meta-Learning (MAML) \cite{finn2017maml}.

The main concept of MAML is to find the meta-model parameters $\bm{\theta}$ which can be adapted to a new task using a small number of samples and gradient steps. This means, it maximizes the sensitivity of the meta-model to changes in the task. In few-shot learning, tasks are defined as a set of new and unknown classes. During training, multiple tasks $\Task_i$ are sampled from a distribution of tasks $P(\mathcal{T})$ and used to optimize the task-specific parameters $\vphi_i$ using a few gradient steps by minimizing a task-specific loss. This is often called the inner loop. The meta-parameters $\vtheta$ are then optimized in the outer loop such that the adaption to $\vphi_i$  of each new task $\Task_i$ maximizes the performance on that task. This results in an optimization over the gradient steps in inner loops, thus a second order optimization. 

\paragraph{Meta Test-Time Training (MT3)}
In our work, we propose a novel combination of self-supervision and meta-learning to have the capability of adapting the model to unknown distributions during test-time. The combination of self-supervision and meta-learning has shown to be beneficial especially for few-short learning \cite{su2020does}. In this work, a self-supervised and a supervised loss where jointly minimized by a meta-learner. In contrast to simply using joint training \cite{sun2020ttt} or minimizing the sum of loss functions by meta-learning \cite{su2020does}, we propose to train the model such that it directly learns to adapt at test-time without supervision. We therefore train the meta-model, parameterized by $\vtheta$, using a supervised and a slightly modified version of BYOL which are combined with MAML. During testing of a single sample, we start with the final meta-model parameters $\vtheta^*$ and fine-tune them for each unlabeled test image to $\vphi^*$ using solely the self-supervised BYOL-like loss. The adapted model is in turn used for inference of that test image.   

For training the meta-parameters in MT3, we define a batch of images as a task $\Task_i$. The parameters $\vtheta$ are transformed to $\vphi_i$ for each task $\Task_i$ by minimizing the modified BYOL loss using two augmented versions of an unlabeled image. The meta-parameters are optimized such that the prediction of the updated model parameterized by $\vphi_i$ leads to a high performance for task $\Task_i$. The optimization of $\vtheta$ is performed over multiple tasks simultaneously as shown exemplarily in Figure \ref{fig:principle} (a).

During testing, illustrated in Figure \ref{fig:principle} (b), a batch of different augmented views of a single test sample defines a task for which we optimize the task-specific parameters with the BYOL-like loss in a self-supervised fashion using one or several gradient steps. This corresponds to the standard version of test-time training \cite{sun2020ttt}. The online setting of \citeA{sun2020ttt} or \citeA{wang2020tent} is not considered further in our work. The optimized parameters $\vphi^*$ for a single sample are only used for the classification prediction of itself and are discarded afterwards. With this test-time adaption we aim for compensating the performance drop caused by unseen test distribution or distribution shifts.

Our contributions are as follows:
\begin{itemize}
	\item We propose a novel combination of meta-learning and self-supervision which is able to adapt to unseen distribution shifts at test-time without supervision.
	\item We analyze MT3 and show that the combination of meta-learning and BYOL achieves better performance than just joint training. 
	\item Our method MT3 significantly outperforms the state-of-the-art in adapting to unseen test distribution shifts.
\end{itemize}



\section{Method}

\begin{figure}[!t]
	\begin{center}
		\def\svgwidth{1.1\linewidth}
		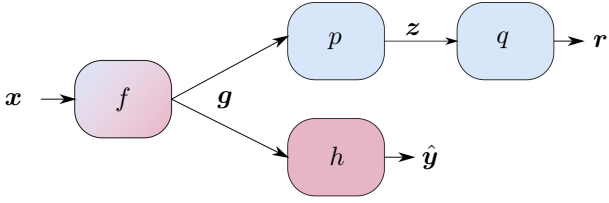
	\end{center}
	\caption{Used architecture in MT3 with shared feature extractor $f$, classification head $h$, and self-supervision head with projector $p$ and predictor $q$.}
	\label{fig:arch}
\end{figure}

\paragraph{Tasks} 
In this work, the training dataset with $N$ input-output pairs is defined as $\DsTrain = \{\vx^k, \vy^k\}_{k=1}^{N}$ with inputs $\vx \in \gX$ and their corresponding class labels $\vy \in \gY$. Each meta-training task $\Task_i$ is associated to a batch of $K$ input-output pairs $\DsTraini= \{\vx_i^k, \vy_i^k\}_{k=1}^{K} $ uniformly sampled from $\DsTrain$. Each update of the meta-parameters is performed over a meta-batch which consists of $T$ tasks. In contrast to MAML, where the meta-objective is the adaption to new tasks, our meta-objective is to adapt the model to unknown data distribution shifts. Therefore, we do not sample the tasks $\Task_i$ with respect to a different set of classes, but different distributions, which is further described in Section \ref{sec:impl}. During testing, a task $\Task$ is defined as adapting the meta-model on $K_T$ augmented views of a single test sample $\vx_{\text{test}}$.   
\paragraph{Architecture}
Similar to previous work in representation learning \cite{Chen2020Simclr,Chen2020_simclrv2,grill2020bootstrap}, the overall architecture as shown in Figure \ref{fig:arch} consists of a feature extractor $f$, a classification head $h$ for a supervised classification, a projector $p$ and a predictor $q$ for an auxiliary self-supervised task. The shared representation $\vg$ will either be used for the classification prediction $\hat{\vy}$ or to calculate the projection $\vz$ and the prediction $\vr$. As introduced by \citeA{Chen2020Simclr,Chen2020_simclrv2}, the similarity in the self-supervised loss is calculated in the projection space $\vz$, instead of the representation space $\vg$. The meta-model is parameterized by the meta-parameters $\vtheta = \left[\vtheta_f, \vtheta_h,\vtheta_p,\vtheta_q,\right]^T$. The task-specific parameters are denoted as $\vphi = \left[\vphi_f, \vphi_h,\vphi_p,\vphi_q,\right]^T $.


\subsection{Meta Test-Time Training}
Since the final model parameters $\vtheta^*$ are adapted at test-time using a single unlabeled test sample  $\vx_{\text{test}}$, the sample-specific parameters $\vphi^*$ are then used for prediction. This test procedure breaks the classical learning paradigm where the model parameters are fixed at test-time. 

In order to make the model adaptable at test-time, there are two crucial problems which need to be addressed. First, we need an unsupervised loss function which is used to adapt the model parameters at test-time, and second, the model has to be optimized during training such that adaptation during test-time results in a better classification performance. 

Following \citeA{sun2020ttt}, a self-supervised loss is minimized in order to update the model parameters at test-time. We make use of BYOL \cite{grill2020bootstrap}, since the rather simple self-supervised rotation loss used by \citeA{sun2020ttt} can fail to provide enough information for adapting the model, specifically if the input sample is rotation invariant. 
\citeA{sun2020ttt} further proposed to jointly minimize a supervised and a self-supervised loss during training. 
Although the authors have shown a correlation between both loss functions under strong assumptions, joint training may not lead to a quickly adaptable model for different self-supervised loss functions as will be shown in our experiments.

In contrast to this, we propose a novel training procedure to directly train a model such that it learns to adapt to unseen samples. 

\paragraph{Meta-Training}
The goal of the meta-training phase is to find the meta-parameters $\vtheta$ which are quickly adaptable to different unseen samples at test-time for achieving a more accurate classification under unknown distribution shifts.  

During one meta-training outer loop step, the minimization of the self-supervised loss $\mathcal{L}_{\text{BYOL}}^{i,k}$ leads to the task-specific parameters $\vphi_i$ for each task $\Task_i$ in the inner loop. The meta-parameters $\vtheta$ are then optimized such that the optimization step to the task-specific parameters leads to high classification accuracy on these $T$ tasks $\Task_i$. 

For each task $\Task_i$, two augmentations $\va_i^k, \tilde{\va}_i^k$ are generated from $\vx_i^k$ using the sample augmentation $\mathcal{A}$ in order to calculate a variation of the BYOL loss as explained in detail in Section \ref{sec:impl}. To further enlarge the differences between the training tasks $\Task_i$, a random batch augmentation $\mathcal{B}$ is applied to $\vx_i, \va_i$ and $\tilde{\va}_i$. Note that the parameters of $\mathcal{B}$ are fixed for all $K$ images within one task and differ across tasks. Therefore, $\mathcal{B}$ artificially generates a distribution shift between tasks and and facilitates meta-learning. 

To calculate our modified BYOL-like loss for each pair $\va_i^k$ and $\tilde{\va}_i^k$, the predictions $\vr_{\vphi_i}^k$ and $\tilde{\vr}_{\vphi_i}^k$ are calculated by the task-specific model parameterized by $\vphi_i$ and the projections $\vz_{\vtheta}^k$ and $\tilde{\vz}_{\vtheta}^k$ using the meta-model $\vtheta$. This differs from the original idea of BYOL where the target model is parameterized by an exponential moving average (EMA) of the online model parameters. In our approach, the meta-model model can be regarded as a smooth version of our task-specific models and therefore a separate target model is obsolete.
Our modified BYOL loss for optimizing the task-specific model is defined as
\begin{equation}
\mathcal{L}_{\text{BYOL}}^{i,k} = \bar{\mathcal{L}}(\vr_{\vphi_i}^k, \tilde{\vz}_{\vtheta}^k) + \bar{\mathcal{L}}(\tilde{\vr}_{\vphi_i}^k, \vz_{\vtheta}^k),
\label{eq:LossBYOL}
\end{equation}
\begin{equation}
\text{with} \quad  \bar{\mathcal{L}}(\va,\vb) =  2 - 2 \cdot \frac{\va^T \vb}{\lVert \va \rVert_2 \cdot \lVert \vb \rVert_2} \nonumber
\end{equation} 
denoting the squared $l_2$-norm of the difference between $l_2$-normalized versions of two vectors $\va$ and $\vb$. The first loss term at the right hand side of Eq. \ref{eq:LossBYOL} measures the closeness of the prediction $\vr_{\vphi_i}^k$ of the task-specific model to the projection $\tilde{\vz}_{\vtheta}^k$ of the meta-model. The second loss term symmetrizes the first one. Note that this loss is only differentiated with respect to the task-specific model parameters $\vphi_i$ excluding the classification head parameters $\vphi_h$. Hence, the $M$ update steps with the inner learning rate $\alpha$ are performed by
\begin{equation}
\vphi_i \leftarrow \vphi_i - \alpha \nabla_{\vphi_i} \frac{1}{K} \sum_{k}\mathcal{L}_{\text{BYOL}}^{i,k}, 
\end{equation} 
where $\vphi_i$ is initialized with the meta-parameters $\vtheta$. 

Now, making use of all optimized task-specific parameters within a meta-batch, the classification predictions for each task $\hat{\vy}_{\vphi_i}^k$ are calculated by the task-specific models parameterized by $\vphi_i$ and, in combination with $\vy_i^k$, are used to optimize the meta-parameters by minimizing the cross-entropy loss $\mathcal{L}_{\text{CE}}(\hat{\vy}_{\vphi_i}^k,\vy_i^k)$. Additionally, the BYOL-like loss function weighted by $\gamma$ is minimized here 
since $\mathcal{L}_{\text{CE}}$ is not differentiable with respect to the parameter of the predictor $p$ and the projector $q$.
The total loss function in the outer loop is defined as
\begin{equation}
\mathcal{L}_{\text{total}}^{i,k} = \mathcal{L}_{\text{CE}}^{i,k} 
+ \gamma \cdot \mathcal{L}_{\text{BYOL}}^{i,k}
\label{eq:LossOut}
\end{equation}
and is calculated using the task-specific parameters $\vphi_i$. 
The update of the meta-parameters $\vtheta$ is done by
\begin{equation}
\vtheta \leftarrow \vtheta - \beta \nabla_\vtheta \frac{1}{KT}\sum_{i} \sum_{k} \mathcal{L}_{\text{total}}^{i,k},
\end{equation}
where $\beta $ is the meta-learning rate. Note that the meta-gradient $\nabla_{\vtheta}$ is a gradient over the optimization steps from $\vtheta$ to every $\vphi_i$. The pseudo-code of the meta-training procedure is described in Algorithm \ref{alg:train}.

\paragraph{Meta-Testing}
At test-time, the optimized meta-model parameters $\vtheta^*$ are adapted to a single test sample $\vx_{\text{test}}$ using the self-supervised BYOL loss in Equation \ref{eq:LossBYOL}. Since only one sample is available during testing, an artificial batch is generated using $K_T$ different augmentation pairs of $\vx_{\text{test}}$ by using the sample augmentation $\mathcal{A}$ to minimize the BYOL-like loss. Using the adapted model, the final classification prediction $\hat{\vy}_{\text{test}}$ is performed. After a prediction, the adapted parameters $\vphi^*$ are discarded and we return back to the final meta-model parameters $\vtheta^*$ for the next test sample. The pseudo-code for processing a single test sample is illustrated in Algorithm \ref{alg:test}.

\begin{algorithm}[!t]
	\caption{Meta-Training} 
	\label{alg:train}
	\begin{algorithmic}[1]
		\State \textbf{Require:} Training data $\DsTrain$, number of inner steps $M$, meta-batch size $T$, task batch size $K$, meta-learning rate $\beta$, inner learning rate $\alpha$, loss weight $\gamma$, sample augmentation $\mathcal{A}$, batch augmentation $\mathcal{B}$
		\While {$\text{not converged}$}
		\State Sample $T$ tasks $\Task_i$ each with batch size $K$
		\For {\textbf{each} $\Task_i$}
		\State Get augmented images $\va_i^k$ and $\tilde{\va}_i^k$ with $\mathcal{A}(\vx_i^k)$
		\State Apply batch augmentation $\mathcal{B}$ to $\vx_i^k,\va_i^k, \tilde{\va}_i^k$
		\For {$step=1,2,\ldots,M$}
		\State Calculate $\vr_{\vphi_i}^k,\tilde{\vr}_{\vphi_i}^k,\vz_{\vtheta}^k,\tilde{\vz}_{\vtheta}^k$
		\State Optimize task-specific model  parameters:
		\State $\vphi_i \leftarrow \vphi_i - \alpha \nabla_{\vphi_i} \frac{1}{K}\sum_{k}\mathcal{L}_{\text{BYOL}}^{i,k} $ \Comment{Eq. \ref{eq:LossBYOL}}
		\EndFor
		\EndFor
		\State Update meta-parameters:
		\State $\vtheta \leftarrow \vtheta - \beta \nabla_\vtheta \frac{1}{KT}\sum_{i} \sum_{k} \mathcal{L}_{\text{total}}^{i,k}$ \Comment{Eq. \ref{eq:LossOut}}
		\EndWhile
		\State \textbf{Return: $\vtheta^*$}
	\end{algorithmic} 
\end{algorithm}
\begin{algorithm}[!t]
	\caption{Test-Time Adaption} 
	\label{alg:test}
	\begin{algorithmic}[1]
		\State \textbf{Require:} Meta-model $\vtheta^*$, test sample $\vx_{\text{test}}$, number of steps $M$, test batch size $K_T$, learning rate $\alpha$, sample augmentation $\mathcal{A}$
		\For {$step=1,2,\ldots,M$}
		\State Initialize $\vphi$ with $\vtheta^*$ 
		\State Copy $\vx_{\text{test}}$ $K_T$ times 
		\State Get $K_T$ different pairs of augmentations $(\va^k, \tilde{\va}^k)$  
		\Statex \hspace{1.2em} from $\vx_{\text{test}}$ with $\mathcal{A}$ 
		\State Optimize task-specific model  parameters:
		\State $\vphi \leftarrow \vphi - \alpha \nabla_{\vphi} \frac{1}{K_T}\sum_{k}\mathcal{L}_{\text{BYOL}}^{k} $ \Comment{Eq. \ref{eq:LossBYOL}}
		\EndFor
		\State Get final classification prediction $\hat{\vy}_{\text{test}}$ 
	\end{algorithmic} 
	
\end{algorithm}


\subsection{Implementation Details}
\label{sec:impl}
\paragraph{Architecture} 
We use a ResNet architecture \cite{he2016resnet} with 26 layers as our feature extractor $f$ with 32 initial filters for all of our experiments. Although the original implementation uses batch normalization (BN), we use group normalization (GN) \cite{wu2018group} with 16 groups  similar to \citeA{sun2020ttt}. 
The projector $p$ and predictor $q$ are each a two-layer MLP with 256 hidden neurons and output dimension of 128. The classifier $h$ shares the first hidden layer with the projector $p$ as proposed by \citeA{Chen2020_simclrv2} followed by a 10-dimensional softmax activated output layer. We empirically found that using no GN in the projector and predictor improves performance. 
\paragraph{Augmentations} For the BYOL-like loss in Equation \ref{eq:LossBYOL}, the sample augmentation $\mathcal{A}$ generates two augmentations of one image. Similar to \citeA{Chen2020Simclr,Chen2020_simclrv2,grill2020bootstrap}, we adjusted the random cropping for CIFAR-10 (uniform between 20 and 32 pixels) and resize back to the original image size of $32 \times 32$. In contrast to other approaches, we apply random vertical flipping with a probability of $\unit[50]{\%}$, since horizontal flipping is already used in the batch augmentation $\mathcal{B}$ and could be reversed if it is applied twice. Lastly, color jittering and color dropping are applied. We use the same types of color jittering as in \cite{grill2020bootstrap} with the adapted strength of $0.2$ compared to $1.0$ for ImageNet \cite{Chen2020_simclrv2}. The color jittering is applied with a probability of $\unit[80]{\%}$ and color dropping with a probability of $\unit[20]{\%}$.   

Additionally, to simulate larger distribution shifts between tasks $\Task_i$ during meta-training, batch augmentation $\mathcal{B}$ is applied to the complete batch $\{\vx_i^k, \va_i^k, \tilde{\va}_i^k\}_{k=1}^K$. The parameters of $\mathcal{B}$ are randomly chosen for each task, but fixed for each image within the current task. Random horizontal flipping ($\unit[50]{\%}$ probability), Gaussian blurring ($\unit[20]{\%}$ probability) with a $3 \times 3$ filter with a standard deviation of $1.0$, brightness adjustment (uniformly distributed delta between $-0.2$ and $0.2$) and Gaussian noise with a uniformly distributed standard deviation between $0$ and $0.02$ are applied. 

\paragraph{Optimization}
We use SGD for the meta-optimization with a fixed learning rate of $\beta=0.01$ and a momentum of $0.9$. The inner optimization is done using only one ($M=1$) gradient step with a step size of  $\alpha=0.1$. During testing, we use the same fixed learning rate of $\alpha=0.1$ and one gradient step since the same parameters are used during training. Weight decay is applied to the meta-model parameters with a strength of $1.5\cdot 10^{-6}$. We set the weight of the BYOL loss to $\gamma=0.1$. Gradient $l_2$-norm clipping with a clipping norm of $10$ is applied to both the inner- and meta-gradient to stabilize the training \cite{finn2017maml}. The meta-batch size is set to $T=4$ and each task consists of $K=8$ images. During test-time adaption, the batch size is set to $K_T=32$. In all experiments the meta-model is trained for 200 epochs which takes approximately 48 hours on a single RTX 2080 Ti (11 GB). Note that the hyper-parameters are only chosen such that the training loss converges. No extensive hyper-parameter optimization was performed.

\paragraph{Dataset}
For training, the CIFAR-10 training dataset \cite{krizhevsky2009learning} is used. For evaluating the test-time training, we use the CIFAR-10-Corrupted dataset \cite{hendrycks2019benchmarking}. It consists of all 10,000 CIFAR-10 validation images with 15 different types of simulated corruptions for 5 different levels. All our results are reported for the most severe level 5. An example image with different corruptions is shown in Figure \ref{fig:corr}. The corruption types come from the four major categories noise, blur, weather, and digital. Exemplary subcategories are impulse noise, Gaussian  blurring, frost, and JPEG compression.

\section{Experiments}
In our experiments, we first analyze the training behavior of our method MT3 followed by a detailed analysis of the impact of meta-learning. For this, we compare to our own baseline and pure joint training (JT) without the meta-learning component. Finally, we compare our results with the state-of-the art method TTT \cite{sun2020ttt}. An overview of all methods is given in Table \ref{tab:methods2}.
\subsection{Ablation Studies}  
\begin{figure}[!t]
	\centering
	\resizebox{\linewidth}{!}{
		\input{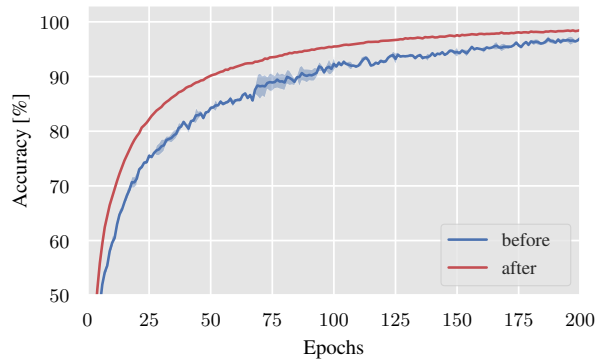}}
	\caption{Training accuracy of tasks $\Task_i$ before and after adaption to task-specific model parameters $\vphi_i$ over 3 runs.}
	\label{fig:plottrainingstd}
\end{figure}

\paragraph{Convergence of MT3} To show its stability and the ability of adaption, we evaluate the classification accuracy during training twice. First, we measure the classification accuracy of each task $\Task_i$ with the meta-model parameters $\vtheta$ before applying the self-supervised adaption. Second, we evaluate the model with the task-specific parameters $\vphi_i$ after the adaption in the inner loop. As shown in Figure \ref{fig:plottrainingstd}, the training of MT3 leads to a stable convergence without large deviations. The small deviations, especially after adaption, highlight the reproducibility and stability of MT3. Furthermore, even at an early stage of training, MT3 learns to adapt such that the accuracy increases as shown by the large gap before and after the self-supervised adaption. This clearly shows that the learned meta-parameters $\vtheta^*$ are able to be adapted with a single gradient step and image resulting in an improved classification accuracy. 
\begin{table}[!t]
	\begin{center}
		\small
		\resizebox{\linewidth}{!}{
			\begin{tabular}{c| c c c c  }
				\toprule
				Method & self-superv. &  $\mathcal{B}$ & training  & test-time adaption \\
				\midrule
				Baseline \cite{sun2020ttt} & -&   - & CE &  -  \\
				JT \cite{sun2020ttt}  & rotation &-  & joint & -\\ 
				TTT \cite{sun2020ttt} & rotation & - & joint & $\checkmark$\\
				\midrule
				Baseline (ours) & -&- & CE & -\\
				JT (ours)  & BYOL &  $\checkmark$ & joint & -\\
				TTT (ours) & BYOL & $\checkmark$ & joint& $\checkmark$\\
				MT (ours)& BYOL &  $\checkmark$& meta& -\\
				MT3 (ours)  &  BYOL & $\checkmark$ & meta & $\checkmark$ \\
				
				\bottomrule
			\end{tabular}
		}
	\end{center}
	\caption{Component overview of all considered methods: Baseline, joint training (JT) and test-time training (TTT) with either cross-entropy (CE), rotation augmentation \cite{gidaris2018rot} or BYOL \cite{grill2020bootstrap}; usage of batch augmentation $\mathcal{B}$; and used training principles, joint training or meta training  with or without test-time adaption.}
	\label{tab:methods2}
\end{table}
\begin{table*}[!t]
	\begin{center}
		\small
		\resizebox{1\linewidth}{!}{
		\begin{tabular}{l | c c c |c c c |c c} 
			\toprule
			& Baseline  & JT & TTT & Baseline  & JT  &  TTT  & MT  & MT3 \\
			&   \cite{sun2020ttt}  &  \cite{sun2020ttt} &  \cite{sun2020ttt} & (ours) &  (ours) &  (ours) & (ours) & (ours)  \\
\midrule
brit & $86.5 $ & $87.4 $ & $\mathbf{ 87.8 }$ & $86.7 \pm 0.44$ & $86.5 \pm 0.13$ & $86.6 \pm 0.26$ &$84.3 \pm 1.15$ &$86.2 \pm 0.47$ \\
contr & $75.0 $ & $74.7 $ & $76.1 $ & $54.0 \pm 6.42$ & $75.4 \pm 2.02$ & $75.1 \pm 2.38$ &$69.3 \pm 2.63$ &$\mathbf{ 77.6 \pm 1.21}$ \\
defoc & $76.3 $ & $75.8 $ & $78.2 $ & $68.1 \pm 2.34$ & $84.7 \pm 0.11$ & $\mathbf{ 84.7 \pm 0.09}$ &$82.7 \pm 1.33$ &$84.4 \pm 0.44$ \\
elast & $72.6 $ & $76.0 $ & $\mathbf{ 77.4 }$ & $74.3 \pm 0.27$ & $74.6 \pm 0.80$ & $74.4 \pm 1.19$ &$74.2 \pm 1.08$ &$76.3 \pm 1.18$ \\
fog & $71.9 $ & $72.5 $ & $74.9 $ & $70.7 \pm 0.98$ & $70.3 \pm 0.86$ & $70.4 \pm 0.67$ &$72.0 \pm 1.03$ &$\mathbf{ 75.9 \pm 1.26}$ \\
frost & $65.6 $ & $67.5 $ & $70.0 $ & $65.2 \pm 0.93$ & $79.8 \pm 0.62$ & $79.5 \pm 0.73$ &$76.6 \pm 1.16$ &$\mathbf{ 81.2 \pm 0.20}$ \\
gauss & $49.5 $ & $50.6 $ & $54.4 $ & $49.9 \pm 3.17$ & $\mathbf{ 71.7 \pm 1.13}$ & $70.4 \pm 1.08$ &$63.6 \pm 1.17$ &$69.9 \pm 0.34$ \\
glass & $48.3 $ & $51.5 $ & $53.9 $ & $50.7 \pm 2.96$ & $62.8 \pm 0.97$ & $61.9 \pm 1.10$ &$62.8 \pm 1.35$ &$\mathbf{ 66.3 \pm 1.24}$ \\
impul & $43.9 $ & $46.6 $ & $50.0 $ & $43.4 \pm 4.31$ & $\mathbf{ 59.3 \pm 3.04}$ & $58.5 \pm 3.17$ &$50.3 \pm 1.68$ &$58.2 \pm 1.25$ \\
jpeg & $70.2 $ & $71.3 $ & $72.8 $ & $76.0 \pm 0.86$ & $78.6 \pm 0.37$ & $\mathbf{ 79.0 \pm 0.44}$ &$75.2 \pm 0.06$ &$77.3 \pm 0.26$ \\
motn & $75.7 $ & $75.2 $ & $77.0 $ & $71.6 \pm 0.46$ & $70.7 \pm 0.45$ & $69.8 \pm 0.46$ &$72.6 \pm 3.17$ &$\mathbf{ 77.2 \pm 2.37}$ \\
pixel & $44.2 $ & $48.4 $ & $52.8 $ & $60.1 \pm 2.73$ & $65.0 \pm 0.32$ & $62.1 \pm 0.44$ &$67.8 \pm 5.13$ &$\mathbf{ 72.4 \pm 2.29}$ \\
shot & $52.8 $ & $54.7 $ & $58.2 $ & $52.3 \pm 2.17$ & $\mathbf{ 72.3 \pm 1.36}$ & $71.0 \pm 1.09$ &$64.0 \pm 1.24$ &$70.5 \pm 0.72$ \\
snow & $74.4 $ & $75.0 $ & $76.1 $ & $74.5 \pm 0.46$ & $77.2 \pm 0.58$ & $77.2 \pm 0.55$ &$77.1 \pm 0.51$ &$\mathbf{ 79.8 \pm 0.63}$ \\
zoom & $73.7 $ & $73.6 $ & $76.1 $ & $67.4 \pm 1.70$ & $81.6 \pm 0.69$ & $\mathbf{ 81.7 \pm 0.66}$ &$78.7 \pm 1.72$ &$81.3 \pm 0.58$ \\
\midrule
\textbf{avg.} & $65.4$ & $66.7$ & $69.0$ & $64.3 \pm 0.42$ & $74.0 \pm 0.77$ & $73.5 \pm 0.80$ & $71.4 \pm 0.42$ & $\mathbf{75.6 \pm 0.30}$ \\
\bottomrule
		\end{tabular}
		}
	\end{center}
	\caption{Performance on the CIFAR-10-Corrupted dataset for MT3 compared to the results from \cite{sun2020ttt}, including their baseline, joint training with rotation loss (JT), and test-time adaption (TTT). Additionally, we report our results of MT3 without test-time adaption (MT), our baseline, joint training (JT), and test-time adaption (TTT) using BYOL. Mean and standard deviation are reported over 3 runs.}
	\label{tab:compare1}
\end{table*}

\paragraph{Baseline} In order to compare MT3 to classical supervised training, we choose the same architecture as described in Section \ref{sec:impl} without the projector and predictor. This baseline model is simply trained by minimizing the cross-entropy loss. We use SGD with a fixed learning rate of 0.1 and a momentum of 0.9. The strength of weight decay is set to $5\cdot 10^{-4}$. We train the baseline model for 200 epochs with a batch size of 128. We use the standard data augmentation protocol by padding with 4 pixels followed by random cropping to $32 \times 32$ pixels and random horizontal flipping \cite{he2016resnet,lee2015deeply}. The hyper-parameters of the baseline training are optimized independently of other methods in order to have a fair comparison.

\paragraph{Joint Training} To show the improvement caused by meta-learning, we compare MT3 to a second baseline, namely joint training (JT). We use exactly the same architecture as for MT3 and minimize the joint loss function similar to Equation \ref{eq:LossOut} but without any inner step, i.e., without meta-learning. Additionally, we use the exponential moving average of the online model as the target model as originally proposed by \citeA{grill2020bootstrap} with an update momentum of 0.996. The BYOL-like loss is weighted by $\gamma=0.1$. For minimizing the joint loss function, we use SGD with a learning rate of 0.1 and a momentum of 0.9. The strength of weight decay is set to $1.5\cdot 10^{-6}$. We train the model for 200 epochs with a batch size of 128. In order to have a fair comparison and to show the impact of meta-learning in MT3, we use the same data augmentation $\mathcal{A}$ for minimizing the BYOL-like loss. Furthermore, we use the same batch augmentation $\mathcal{B}$ to simulate distributions shifts here as well. The only major difference to MT3 is the use of joint training instead of meta-learning.

On the one hand,  we use joint training to compare it to MT3 by fixing the learned model at test-time. On the other hand, similar to \citeA{sun2020ttt}, we adapt our jointly trained model at test-time using only the self-supervised loss (TTT). During test-time adaption, we use the same test-time parameters as for MT3 except the learning rate is set to 0.01 which is more comparable to the effective learning rate during joint training (due to $\gamma=0.1$). The hyper-parameters of joint training are again optimized independently of other methods.


\begin{figure*}[!t]
	\begin{centering}
		\includegraphics[width=0.9\linewidth]{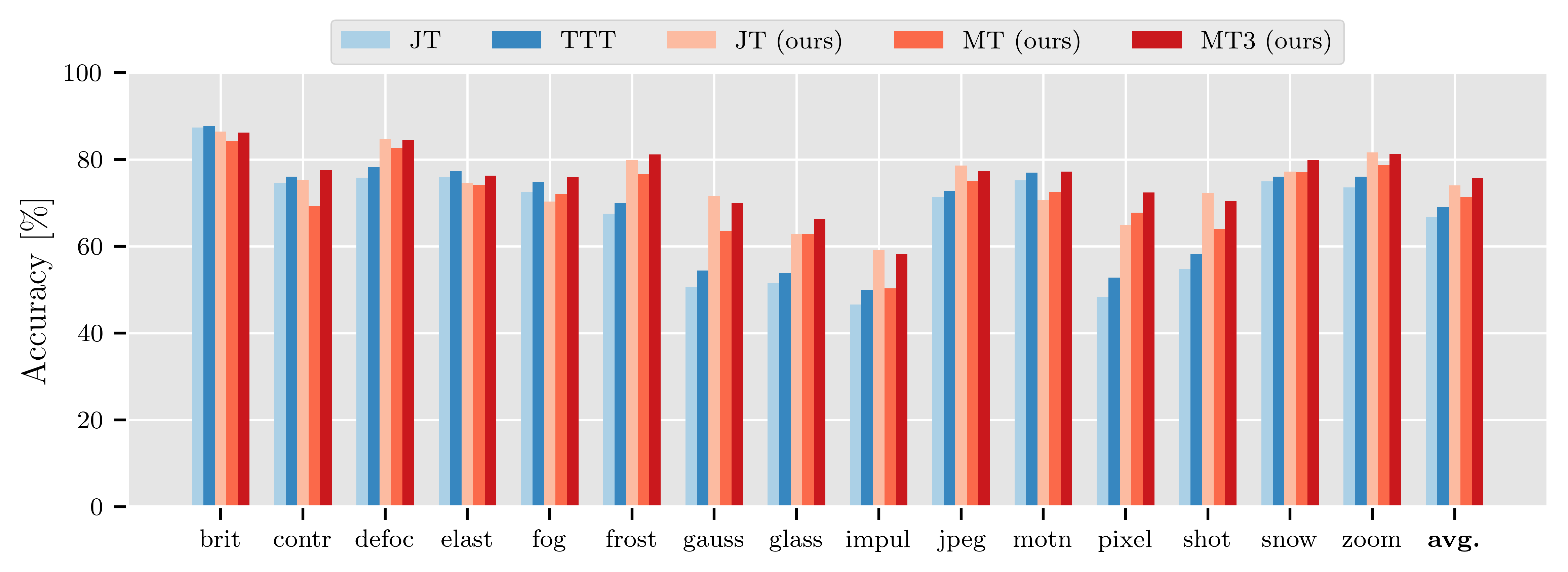}
	\end{centering}
	\caption{Performance on the CIFAR-10-Corrupted dataset for MT3 compared to MT3 without test-time adaption (MT), joint training with BYOL (JT (ours)), joint training with rotation loss (JT) \cite{sun2020ttt}, and test-time training (TTT) \cite{sun2020ttt}. Mean and standard deviation are reported over 3 runs.}
	\label{fig:barplot}
\end{figure*}

\paragraph{Comparison to our baselines} We first compare our baseline and joint training without test-time adaption against our proposed method MT3. Additionally, we show the results of MT3 with fixed parameters at test-time (MT), thus without a gradient step at test-time. The results on the 15 corruption types of the CIFAR-10-Corrupted images are shown in Table \ref{tab:compare1} with their mean and standard deviation estimated over 3 runs. Furthermore, the average accuracy over all corruption types for each run is given by its mean and standard deviation. In case of TTT, the model parameters are adapted before the prediction of each single test image. The final accuracy is then calculated over the predictions of the 10,000 adapted models. 

Our baseline model has on average the worst performance with an accuracy of $\unit[64.3]{\%}$. In comparison, our JT with stronger data augmentation and the utilization of the BYOL loss leads to a $\unit[9.7]{\%}$ increase in accuracy achieving $\unit[74.0]{\%}$. Applying test-time training to our jointly trained model, the average accuracy drops down to  $\unit[73.5]{\%}$, contrary to our expectations. Although joint training followed by test-time training is expected to help improving the result as shown in \cite{sun2020ttt}, we did not experience this in our case, where BYOL instead of a rotation loss is used. For some corruption types, e.g. jpeg compression (jpeg), a small improvement can be achieved with our TTT, but in 9 of 15 cases the test accuracy decreases with test-time training, e.g. for pixelate (pixel) by almost $\unit[3]{\%}$. In contrast, our method MT3 achieves a higher classification accuracy for all types of corruption after performing test-time adaption. MT3 raises the average accuracy of the meta-model from $\unit[71.4]{\%}$ before  to $\unit[75.6]{\%}$ after adaption. Considering the average over all corruptions, MT3 has the lowest standard deviation, which highlights the stability and reproducibility of our method. Similar to JT, the results of the two corruption types Gaussian noise (gauss) and brightness (brit), which overlap with the applied batch augmentation $\mathcal{B}$, have improved compared to our baseline. The improvement on these datasets is mainly caused by the applied data augmentation and should therefore be handled carefully. Still, our method MT3 outperforms JT on average despite both methods using the same data augmentations. 

In summary, the results suggest that our proposed method MT3 has learned during training to adapt at test-time, while joint training using BYOL combined with test-time adaption did not show that behavior. Furthermore, our analysis shows that the absolute improvement of MT3 is caused by meta-training and not only by using joint training with stronger data augmentation.

\subsection{Comparison with state-of-the art}

We compare our method to the state-of-the-art TTT \cite{sun2020ttt} as shown in Table \ref{tab:compare1} and Figure \ref{fig:barplot}. Besides our results, we discuss the baseline, joint training (JT) and joint training with test-time adaption (TTT) of \citeA{sun2020ttt}. The difference between all analyzed methods are shown in Table \ref{tab:methods2}.   

Our baseline as well as the baseline of \citeA{sun2020ttt} have similar average performance over all corruption types. This highlights that both models have a comparable capacity or generalization capability and possible improvements are not caused by the model structure itself. In our work, joint training with the BYOL-like loss leads to a much higher average accuracy compared to the previous method where rotation classification as self-supervision was used. The large gap of $\unit[7.3]{\%}$ might be caused by the stronger data augmentations or the use of BYOL in our method. Despite this, one important result is that simple joint training does not enable the ability to adapt at test-time in general. In the previous work of \citeA{sun2020ttt}, the adaption with only the self-supervised loss to a single test image using ten gradient steps leads on average to an improvement of $\unit[2.3]{\%}$. In comparison, test-time adaption for our jointly trained model using BYOL leads to an average degradation of $\unit[0.5]{\%}$. We also investigated the case of ten gradient steps at test-time, but found that on average the performance further degrades.  

Our method MT3, on the other hand, shows the ability to adapt by a large improvement of $\unit[4.2]{\%}$ before and after a single gradient step. Furthermore, the final average accuracy of $\unit[75.6]{\%}$ over all corruption types is the best among all considered methods. For 7 out of 15 corruption types, MT3 has the highest accuracy compared to our baselines and previous work. This again highlights the ability of our method to adapt to unseen distribution shifts using a single gradient step during test-time.

\begin{table}[!t]
	\begin{center}
		\small
		\resizebox{0.48\textwidth}{!}{
			\begin{tabular}{l | c c c |c c} 
\toprule
&  Baseline & JT & TTT  & MT  & MT3 \\
\midrule
brit & $\mathbf{ 54.5 \pm 0.79}$ & $53.2 \pm 0.75$ & $53.3 \pm 0.79$ &$51.7 \pm 0.42$ &$52.2 \pm 0.44$ \\
contr & $22.2 \pm 1.60$ & $29.0 \pm 2.45$ & $28.2 \pm 2.49$ &$28.7 \pm 0.55$ &$\mathbf{ 31.6 \pm 1.53}$ \\
defoc & $37.9 \pm 2.47$ & $55.8 \pm 0.32$ & $\mathbf{ 55.9 \pm 0.49}$ &$54.8 \pm 0.62$ &$55.0 \pm 0.55$ \\
elast & $\mathbf{ 45.0 \pm 0.67}$ & $44.0 \pm 0.90$ & $44.0 \pm 0.88$ &$44.1 \pm 0.61$ &$44.2 \pm 0.81$ \\
fog & $30.8 \pm 0.94$ & $31.9 \pm 1.31$ & $32.0 \pm 1.35$ &$32.4 \pm 0.27$ &$\mathbf{ 33.3 \pm 0.45}$ \\
frost & $31.3 \pm 0.96$ & $44.7 \pm 0.78$ & $44.5 \pm 0.94$ &$43.8 \pm 0.90$ &$\mathbf{ 45.5 \pm 1.00}$ \\
gauss & $18.7 \pm 2.86$ & $\mathbf{ 32.8 \pm 2.01}$ & $32.1 \pm 1.94$ &$30.7 \pm 1.05$ &$32.8 \pm 0.84$ \\
glass & $25.8 \pm 1.21$ & $30.5 \pm 1.77$ & $30.2 \pm 1.95$ &$31.7 \pm 1.02$ &$\mathbf{ 33.0 \pm 0.93}$ \\
impul & $14.2 \pm 2.33$ & $17.1 \pm 0.19$ & $16.9 \pm 0.26$ &$17.3 \pm 0.21$ &$\mathbf{ 18.4 \pm 0.09}$ \\
jpeg & $\mathbf{ 44.1 \pm 1.51}$ & $43.1 \pm 0.85$ & $43.2 \pm 0.83$ &$42.4 \pm 0.27$ &$42.7 \pm 0.46$ \\
motn & $40.7 \pm 2.08$ & $44.8 \pm 0.71$ & $44.4 \pm 0.74$ &$44.4 \pm 0.94$ &$\mathbf{ 45.4 \pm 0.81}$ \\
pixel & $28.1 \pm 0.67$ & $34.7 \pm 0.59$ & $33.2 \pm 0.79$ &$40.8 \pm 1.86$ &$\mathbf{ 41.2 \pm 2.06}$ \\
shot & $20.0 \pm 2.69$ & $32.9 \pm 1.52$ & $32.2 \pm 1.52$ &$31.1 \pm 1.28$ &$\mathbf{ 33.1 \pm 1.41}$ \\
snow & $38.4 \pm 0.77$ & $42.9 \pm 0.85$ & $42.7 \pm 0.90$ &$43.2 \pm 0.62$ &$\mathbf{ 43.7 \pm 1.12}$ \\
zoom & $39.7 \pm 2.10$ & $53.8 \pm 1.04$ & $\mathbf{ 54.0 \pm 1.04}$ &$52.2 \pm 0.56$ &$52.0 \pm 0.62$ \\
\midrule
\textbf{avg.} & $32.8 \pm 0.49$ & $39.4 \pm 0.42$ & $39.1 \pm 0.40$ & $39.3 \pm 0.37$ & $\mathbf{40.3 \pm 0.27}$ \\
\bottomrule
			\end{tabular}
		}
	\end{center}
	\caption{Accuracy on the CIFAR-100-Corrupted dataset for MT3 compared to our baselines. Mean and standard deviation are reported over 3 runs.}
	\label{tab:compare-cifar100}
\end{table}
\subsection{CIFAR-100-Corrupted}
To show the success and scalability of MT3, we evaluate our method on the more challenging CIFAR-100-Corrupted dataset \cite{krizhevsky2009learning, hendrycks2019benchmarking}. Since \citeA{sun2020ttt} did not evaluate this dataset and \citeA{wang2020tent} only for the online adaption, we only show the results compared to our baselines. We use the same hyperparameter as for CIFAR-10 except the test learning rate is lowered to $\alpha=0.05$. As shown in Table \ref{tab:compare-cifar100}, our method is also capable to learn to adapt on the more complex dataset CIFAR-100 with a similar behavior as for the CIFAR-10-Corrupted dataset. 

\section{Conclusion}
We proposed a novel algorithm that allows to adapt to distribution shifts during test-time using a single sample. We show that our approach, based on meta-learning (MAML) and self-supervision (BYOL), effectively enables adaptability during test-time. In contrast to the previous work, where simply joint training was used, meta-learning has the explicit purpose to learn meta-parameters that can be rapidly adapted which we showed in our experiments. Our combination of meta-learning and self-supervision improves the average accuracy on the challenging CIFAR-10-Corrupted dataset by $6.6\%$, a $9.57\%$ relative increase, compared to the state-of-the-art TTT.

\bibliographystyle{apacite}
\bibliography{egbib}
\end{document}